\documentclass[10pt,twocolumn,letterpaper]{article}

\usepackage[pagenumbers]{cvpr} 

\usepackage[dvipsnames,table]{xcolor} 
\usepackage{afterpage}
\usepackage{booktabs} 
\usepackage{tikz} 
\usetikzlibrary{positioning} 
\usetikzlibrary{arrows} 
\usetikzlibrary{shapes.geometric} 
\usepackage{soul}

\definecolor{cvprblue}{rgb}{0.21,0.49,0.74}
\usepackage[pagebackref,breaklinks,colorlinks,citecolor=cvprblue]{hyperref}

\title{Local Masking Meets Progressive Freezing: Crafting Efficient Vision Transformers for Self-Supervised Learning}

\author{Utku Mert Topcuoglu, Erdem Akagündüz\\
Graduate School of Informatics\\
METU, Ankara, Türkiye\\
{\tt\small \{utku.topcuoglu,akaerdem\}@metu.edu.tr}
}

\begin{document}
\maketitle
\begin{abstract}
In this paper, we present an innovative approach to self-supervised learning for Vision Transformers (ViTs), integrating local masked image modeling with progressive layer freezing. This method focuses on enhancing the efficiency and speed of initial layer training in ViTs. By systematically freezing specific layers at strategic points during training, we reduce computational demands while maintaining or improving learning capabilities. Our approach employs a novel multi-scale reconstruction process that fosters efficient learning in initial layers and enhances semantic comprehension across scales. The results demonstrate a substantial reduction in training time (~12.5\%) with a minimal impact on model accuracy (decrease in top-1 accuracy by 0.6\%). Our method achieves top-1 and top-5 accuracies of 82.6\% and 96.2\%, respectively, underscoring its potential in scenarios where computational resources and time are critical. This work marks an advancement in the field of self-supervised learning for computer vision. The implementation of our approach is available at our project's GitHub repository: \url{https://github.com/utkutpcgl/ViTFreeze}.
\end{abstract}

\section{Introduction}

Training deep learning models poses significant challenges due to their complexity and the substantial computational resources they demand. Vision Transformers (ViT) \cite{dosovitskiy2020image} exemplify this challenge, requiring compute and time to train. Consequently, the research community has been actively developing methods to expedite the training of such models.

Various strategies have been proposed to accelerate model training. Techniques such as network pruning and quantization aim to shorten training duration. Recent works in network pruning from a sparsity perspective have shown the potential to enable rapid AI deployment in devices with computation and memory constraints \cite{diao2020pruning, survey2020pruning, efficacy2020pruning}. On the other hand, advancements in quantization have focused on creating methods that balance accuracy and latency, particularly in edge deployments and post-training scenarios \cite{automated2020quantization, efficient2020quantization, qat2020quantization}. Unique approaches, akin to ours, leverage the self-supervised learning paradigm to improve training efficiency through mechanisms like masked image modeling \cite{xue2023stare, anonymous2023fastmim}.

In this work, we refine the FreezeOut\cite{brock2017freezeout} technique, originally designed to accelerate the training of deep networks by progressively freezing layers, and adapt it to the multi-scale architecture of Vision Transformers engaged in local masked image modeling\cite{wang2023masked}. Our method systematically identifies and freezes specific layers at strategic points during training, reducing computational overhead while maintaining or enhancing the model's learning capability.

Our contributions are thus twofold. \textbf{First}, we propose a novel method that adapts the FreezeOut technique for the hierarchical structure of Vision Transformers, significantly improving training efficiency. \textbf{Second}, we validate our approach through extensive experiments, showcasing its ability to maintain high performance with reduced training time. Moreover, we discuss the potential of our method to scale to larger and more complex models, potentially revolutionizing efficient training practices in self-supervised learning.

\section{Related Works}
\textbf{Self-supervised learning in Vision.}
Self-supervised learning has become a significant trend in computer vision, marking a shift from supervised to autonomous representation learning. Key methodologies include generative techniques like denoising autoencoders \cite{vincent2008extracting}, image inpainting \cite{pathak2016context}, and colorization \cite{larsson2017colorization}, which reconstruct data from partial inputs. Contrastive learning, a form of discriminative approach, is notable for creating varied image views and ensuring consistent representations \cite{chen2020simple, he2020momentum}, sometimes without contrasting negative pairs \cite{grill2020bootstrap, huang2021self}. Recent approaches in contrastive learning also incorporate spatial aspects in their analysis \cite{wang2021dense}, highlighting the versatility and adaptability of self-supervised learning in visual data processing.

\noindent\textbf{Masked Language/Image Modeling}
In the evolving landscape of self-supervised learning, Masked Image Modeling (MIM) and Masked Language Modeling (MLM) have become instrumental. MLM, pioneered by BERT \cite{devlin2019bert} and its variations \cite{brown2020language}, has significantly impacted natural language processing by predicting randomly masked tokens in sentences. Paralleling this, MIM, influenced by the development of Vision Transformers \cite{dosovitskiy2020image}, \cite{bao2021beit}, adopts a similar approach for images, predicting corrupted or masked parts. While early methods like \cite{bao2021beit} and \cite{he2021masked} set the foundation, recent advancements \cite{xie2021simmim}, \cite{wei2021masked} have pushed MIM to the forefront, especially in fine-tuning performance for various visual tasks. These MIM strategies, varying in target signals from normalized pixels \cite{gao2022convmae}, discrete tokens \cite{dong2021peco}, to handcrafted features \cite{wei2021masked}, have overcome the initial limitations of computational intensity and pre-training duration. Innovative approaches such as \cite{huang2022green} and \cite{li2022uniform} optimize the encoding process, and ConvMAE \cite{gao2022convmae} leverages local multi-scale reconstructions, further enhancing the efficiency and adaptability of MIM in diverse applications.

Expanding upon the existing self-supervised learning frameworks, recent developments in efficient Masked Image Modeling (MIM) have notably advanced the field. "MixMAE" by Liu et al. \cite{liu2022mixmae}, introduced in 2022, brought innovative autoencoder designs for vision transformers. The work of Wang et al. \cite{wang2023masked}, "Masked Image Modeling with Local Multi-Scale Reconstruction," further contributes to the field by enhancing the MIM process with a local multi-scale reconstruction approach, providing a comprehensive framework for feature extraction and model robustness. In 2023, "FastMIM" \cite{anonymous2023fastmim} enhanced pre-training efficiency using low-resolution images. Similarly, "Disjoint Masking with Joint Distillation" by Ma et al. \cite{ma2023disjoint} provided a novel approach to distillation. Particularly notable is "Stare at What You See" by Xue et al. \cite{xue2023stare}, which diverges from traditional reconstruction methods in MIM, opting for a unique approach focused on non-reconstruction-based image modeling. Other contributions like "M2T" by Mentzer et al. \cite{mentzer2023m2t} and "Attention-Guided Masked Image Modeling" \cite{wang2023what} further emphasize the trend towards more efficient and applicable MIM strategies. These developments represent significant steps toward refining MIM's efficiency and usability.

\noindent\textbf{Speeding up Training with Layer Freezing and Dropping.} The advancement of efficient deep neural network (DNN) training has been significantly influenced by layer freezing and dropping techniques. Bengio et al. (2006) pioneered this field with their work on layer-wise training, which highlighted the complexity of training deep architectures \cite{bengio2006greedy}. Building on this, Brock et al. (2017) introduced FreezeOut, a method for progressively freezing layers, which showed time savings in training with minimal accuracy loss for certain architectures \cite{brock2017freezeout}. Xiao et al. (2019) furthered this approach by proposing an intelligent layer freezing method based on normalized gradient differences, enhancing training efficiency \cite{xiao2019fast}. Zhang and He (2020) extended these concepts to Transformer-based language models, achieving substantial reductions in training time and computational costs through progressive layer dropping \cite{zhang2020accelerating}. Chen et al. (2022) contributed by identifying a Layer Convergence Bias, indicating faster convergence in shallower layers of DNNs, which supports the idea of selective layer training \cite{chen2022layer}. Wang et al. (2022) introduced a knowledge-guided layer freezing technique, using semantic knowledge from a reference model to accelerate training without sacrificing accuracy \cite{wang2022egeria}. Liu et al. (2021) developed AutoFreeze, an adaptive system for accelerating model fine-tuning by selectively training layers and efficiently caching intermediate activations \cite{liu2021autofreeze}. Finally, Yang et al. (2023) applied these concepts to self-supervised continual learning, using progressive task-correlated layer freezing to further enhance training efficiency \cite{yang2023efficient}. These developments collectively mark a substantial evolution in the acceleration of DNN training, showcasing a strategic approach to managing individual layers’ training processes.
\label{sec:approach}

\section{Approach}

\subsection{Multi-scale Masked Image Modeling}
 
\textbf{Masked Image Modeling.}
Masked image modeling, a self-supervised learning technique, involves randomly masking a subset of input data and training a model to predict these masked portions \cite{he2021masked, bao2021beit, xie2021simmim}. Mathematically, given an input image $x \in \mathbb{R}^{H \times W \times C}$, it is divided into a set of patches $\{x_p^i\}_{i=1}^{N}$, where $N = HW/P^2$ is the number of patches for a patch size $P$. A subset of these patches $\{x_p^m\}_{m=1}^{M}$, where $M < N$, is masked. The goal is to predict the masked patches $\hat{x}_p^m$ based on the unmasked ones, leveraging the contextual information provided by the visible patches. This approach is analogous to the masked language modeling in NLP, as seen in BERT \cite{devlin2019bert}, and enables the model to learn rich, contextual representations of the input data without the need for labeled examples. This method is particularly effective in vision transformers \cite{dosovitskiy2020image}, where the global context captured by the transformer aids in accurately reconstructing the masked patches.

\begin{figure*}[htbp]
  \centering
  \includegraphics[width=0.8\linewidth]{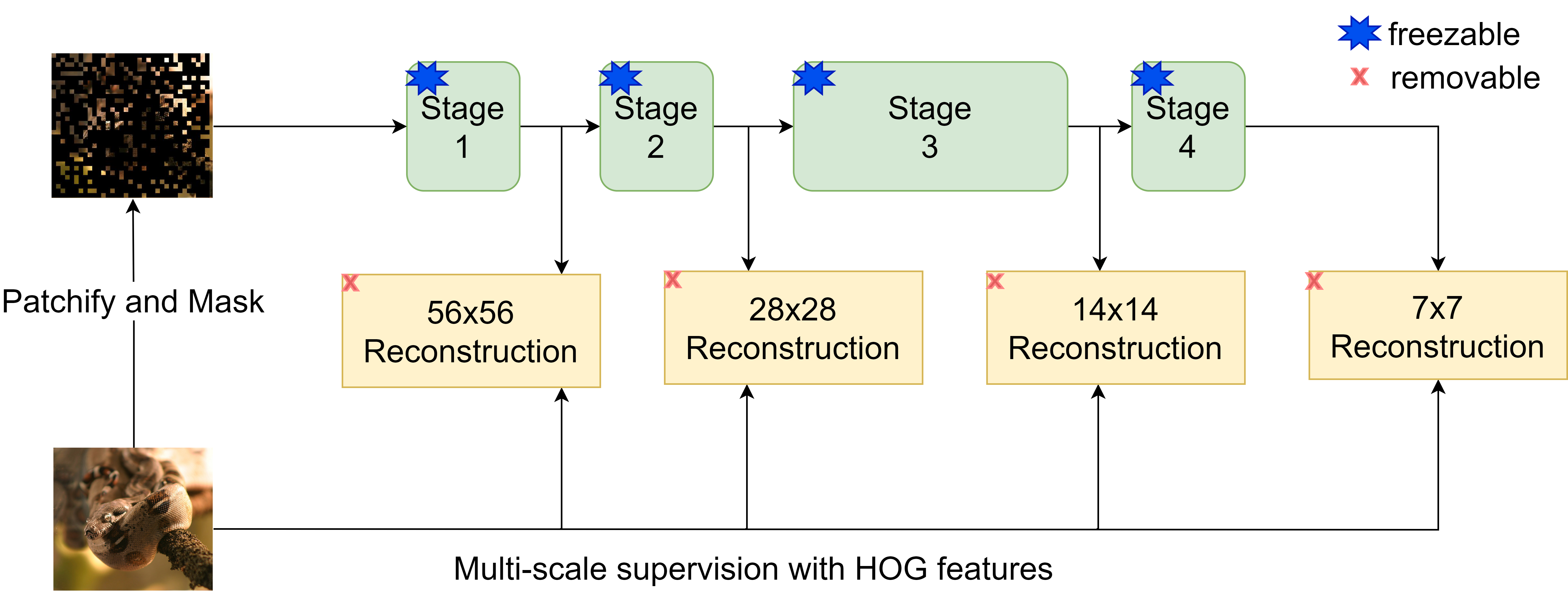}
  \caption{Proposed method with progressive freezing of layers (stages) and removal of decoders and reconstruction tasks once they are no longer necessary. Layers are being frozen individually from the encoder stages, while the reconstruction task (decoder and hog layer) is removed totally when the input encoder stage has been frozen.}
  \label{fig:localmim_freezeout}
\end{figure*}

\noindent\textbf{Local Masked Image Modeling.}
In Masked Image Modeling (MIM), the lower encoder layers are pivotal for pre-training, as they propagate crucial semantical knowledge to the upper layers. However, these initial layers struggle with learning inter-patch semantical relations due to the self-attention mechanism's complexity in vision transformers, especially under global reconstruction loss. For the initial layers to learn better relations between the patches, the multi-scale method has been proposed.

Our approach enhances the learning of initial layers in vision transformers through Local Masked Image Modeling (LocalMIM) \cite{wang2023masked}. This method leverages multi-scale reconstruction to explicitly guide multiple layers in the model, facilitating a nuanced understanding of the input at varying scales. We divide the input image into non-overlapping regions, each providing supervision signals for the reconstruction task at different scales. The details of our proposed approach are depicted in Figure \ref{fig:localmim_freezeout}.

In Local multi-scale reconstruction for Masked Image Modeling (MIM), an image $x \in \mathbb{R}^{H \times W \times C}$ is divided into regions $\{x_i \in \mathbb{R}^{p \times p \times C}\}_{i=1}^{HW/p^2}$. Supervision signals $y_i = \pi(x_i)$ are extracted with the HOG feature descriptor denoted with $\pi$, fine-scale supervisions capturing low-level details and coarse-scale supervisions encapsulating high-level semantics \cite{wang2023masked}.

The model's lower layers are tasked with reconstructing fine-scale supervision, while the upper layers focus on coarse-scale supervision, as it has been observed that applying the coarse-scale reconstruction task to different stages does not improve the model's performance solely and fine-scale to coarse-scale supervision performs the best \cite{wang2023masked}. The decoder, comprising Transformer blocks and Deconvolution/Pool, rescales the predictions to match the supervision scale (in 4 levels).

The training loss is formulated as:
\[
\mathcal {L}_{LocalMIM}=-\sum _{l\in \mathcal {I}}w_l\cdot \sum _{i=1}^{N_l}m_i^l\cdot \ln P(y_i^l|\hat {y}_i^l),
\]
where \(\mathcal {I}\) represents the chosen layers, \(w_l\) the weight of each local loss, and \(m_l\) the mask, where the losses are weighted equally in LocalMIM. This method effectively leverages multi-scale supervision, enhancing the learning process and semantic understanding across different scales.

\noindent\textbf{Vision Transformers.}
In Vision Transformers (ViT) for masked image modeling, an image $x \in \mathbb{R}^{H \times W \times C}$ is transformed into a sequence of $N$ 2D patches $x_p \in \mathbb{R}^{N \times (P^2 \cdot C)}$, where $N = HW/P^2$ and $P$ is the patch size.
 These patches are projected to $D$ dimensions, forming patch embeddings. A learnable (class) token is added to the sequence, and 1D positional embeddings are incorporated to retain spatial context. The Transformer encoder processes the sequence, comprising multi-headed self-attention (MSA) and linear blocks with layer normalization (LN) and residual connections. For masked image modeling, a portion of the patches is randomly masked, and the model is trained to reconstruct these patches, thus learning contextualized representations. This methodology, inspired by the principles of masked language modeling, enables the ViT to capture intricate visual patterns and long-range dependencies within the image data.

\subsection{Progressive Layer Freezing}
FreezeOut introduces a modification to the backpropagation and Stochastic Gradient Descent (SGD) pipeline, aimed at reducing training time \cite{brock2017freezeout}. This method employs a layer-wise cosine annealing schedule, where the learning rate of each layer $L_i$ is gradually reduced to zero. Specifically, the learning rate $\alpha_i(t)$ of layer $i$ at iteration $t$ follows the equation:
\begin{equation}
    \alpha_i(t) = 0.5 \times \alpha_i(0) \times (1 + \cos(\frac{\pi t}{t_i}))
\end{equation}
Here, $\alpha_i(0)$ represents the initial learning rate of layer $i$, and $t_i$ is the iteration at which its learning rate is scheduled to reach zero. The freezing out of layers begins at a predefined iteration $t_0$, with subsequent layers' learning rates reaching zero at progressively later iterations. Once a layer's learning rate hits zero, it transitions to inference mode and is excluded from all future backward passes. This process results in a per-iteration speedup, proportional to the computational cost of the frozen layer. Additionally, FreezeOut allows for variations in the scaling of initial layer-wise learning rates and the relation of $t_0$ to the freezing times of other layers, enabling flexibility in prioritizing the training of later layers.

In FreezeOut, two aspects of the training strategy are varied experimentally. Firstly, the initial layer-wise learning rate \(\alpha_i(0)\) is scaled relative to the base learning rate \(\alpha\) and the time \(t_i\) at which the layer's learning rate reaches zero. This scaling is defined as:
\begin{equation}
    \alpha_i(0) = \frac{\alpha}{t_i}
\end{equation}
where \(\alpha\) is the base learning rate, also used for the final layer the time \(t_i\) is valued between 0-1 (1 representing the whole training). This scaling ensures that each layer's learning curve integrates to the same value, allowing each layer to traverse an equivalent distance in the weight space, despite a reduced number of training steps.

Secondly, the relationship between the initial freezing time \( t_0 \) and the freezing times \( t_i \) for the remaining layers is varied. A straightforward approach adopts a linear scheduling rule, with the \( t_i \) values being cubed, denoted as \( t_{i,\text{cubed}} = t_{i,\text{linear}}^3 \). This adjustment gives more training priority to later layers compared to a linear schedule. They explore both an unscaled variant, where all \( \alpha_i \) values are identical, and a scaled variant, where the \( \alpha_i \) values are scaled based on the cubed \( t_i \) values. For instance, a user-selected \( t_0 = 0.5 \) would result in a cubed \( t_{0,\text{cubed}} = 0.125 \). These modifications to \( t_0 \) and the FreezeOut strategy introduce flexibility to the training process, potentially reducing the need for extensive hyperparameter tuning \cite{brock2017freezeout}.

\subsection{ViT Freeze}
Our approach aims to enhance the learning efficiency and speed of initial layers in Vision Transformer (ViT) based masked image modeling. By integrating local reconstruction tasks, these layers gain a deeper understanding of the input data at an accelerated pace. This strategy allows for earlier freezing of the layers while minimizing any loss in model accuracy. This synergy between advanced local masked image modeling and progressive layer freezing is designed to optimize both the learning quality and computational efficiency of ViTs.

In our model, we incorporate a transformer block design that is amenable to freezing. Specifically, in the ViT encoder model, every layer, starting from the patch embedding layer, is progressively frozen during training. In the ViT-B model, for instance, there are 13 such layers, 12 transformer blocks, and one patch embedding layer. When a layer is frozen, all of its parameters, including weights and biases, are concurrently transitioned to a non-trainable state.

We selected a \( t_0 \) value of 0.8, followed by cubic scheduling and learning rate scaling, as our default strategy as it was proposed in Freezeout to be the lowest value with a low accuracy drop. The cubic scheduling approach, especially with the cubed \( t_0 \) value of 0.5120 (from the original \( t_0 = 0.8 \)), prioritizes the training of later layers slightly more effectively compared to linear scheduling.

\begin{figure}[t]
  \centering
  \includegraphics[width=0.8\linewidth]{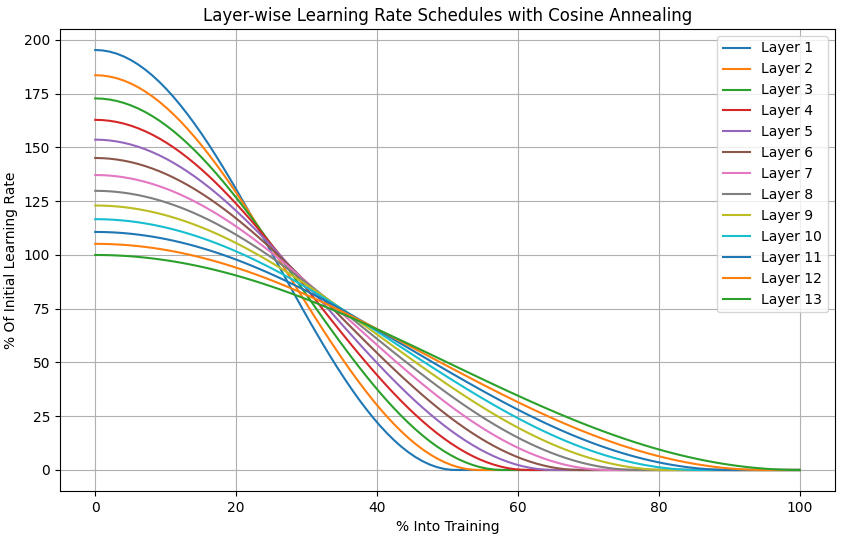}
  \caption{13 layer cubic learning rate scheduling with \( t_0 = 0.8 \).}
  \label{fig:lr_schedule}
\end{figure}

Applying Freezeout logic to ViT-B is demonstrated in Figure~\ref{fig:lr_schedule}. The learnable patch embedding layer is taken as the first layer in addition to the 12 leading transformer encoder blocks.

\begin{figure}[t]
  \centering
  \includegraphics[width=0.8\linewidth]{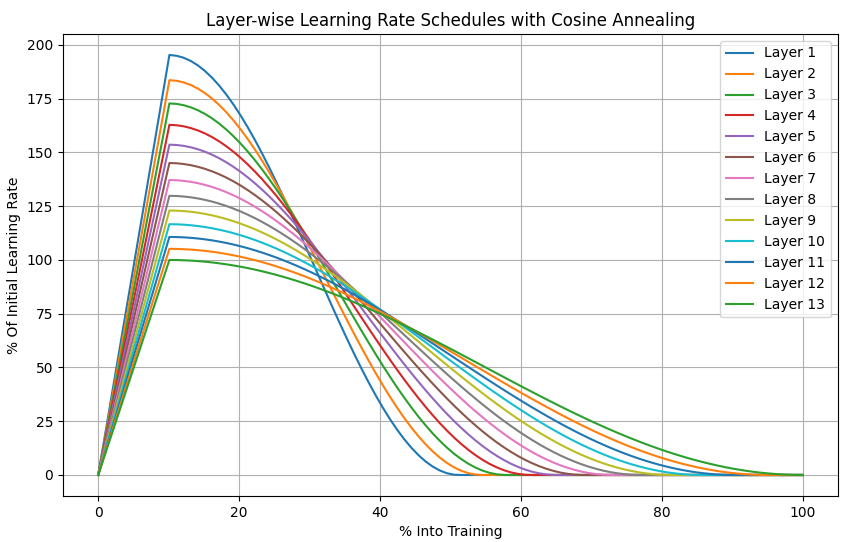}
  \caption{13 layer cubic learning rate scheduling for \( t_0 = 0.8 \) with 10\% warm-up.}
  \label{fig:lr_schedule_with_warmup}
\end{figure}

Our method rather linearly warm-ups the learning rate of layers to their initial learning rate as shown in  Figure~\ref{fig:lr_schedule_with_warmup}. Warm-up with the Adam optimizer gradually increases the learning rate from a low to the target rate, enhancing initial training stability. This approach prevents large, destabilizing weight updates at the start, allowing for smoother and more effective convergence during the early optimization steps

\textbf{Pruning Decoders}
Our method further improves the speed by dynamically removing intermediate decoders from the calculation as soon as the encoder feeding them is frozen. By selectively removing these decoders, we prevent unnecessary computations, thereby enhancing the efficiency of the training process further.

\label{sec:experiments}


\begin{table*}[t]
\centering
\caption{Comparing fine-tuning accuracy on ImageNet-1K with LocalMIM ViT-B 100 epoch pre-training. Bold means faster.}
\label{our-table}
\begin{tabular}{lccccccc}
\hline
\textbf{Model} & \textbf{Backbone} & \textbf{\# Params} & \textbf{PT Epoch} & \textbf{GPU Hours/Ep.} & \textbf{Total GPU Hours} & \textbf{Top-1} & \textbf{Top-5} \\ \hline
LocalMIM-HOG & ViT-B & 86M & 100 & 0.48 & 48 & 83.2 & 96.6\\
Ours & ViT-B & 86M & 100 & \textbf{0.42} & \textbf{42} & 82.5 & 96.2 \\ \hline
\end{tabular}
\end{table*}

\begin{table*}[t]
\centering
\caption{Comparing fine-tuning accuracy on ImageNet-1K as reported in LocalMIM\cite{wang2023masked}. The models underwent pre-training and fine-tuning at a resolution of 224 x 224, with the exception of SimMIM$_{192}$, which was pre-trained at a resolution of 192 x 192. The symbol {\textdagger} denotes the employment of relative positional encoding. Our experiment was hypothetically extended to the LocalMIM evaluation setup by keeping the speed-up ratio for a broader and more fair comparison across different methods. Bold means faster.}
\label{localmim-table}
\begin{tabular}{lcccccc}
\hline
\textbf{Model} & \textbf{Backbone} & \textbf{\# Params} & \textbf{PT Epoch} & \textbf{GPU Hours/Ep.} & \textbf{Total GPU Hours} & \textbf{Acc} \\ \hline
Scratch, ViT & ViT-B & 86M & 0 & 1.5 & - & 82.3 \\
MoCo v3 & ViT-B & 86M & 600 & - & - & 83.2 \\
DINO & ViT-B & 86M & 300 & - & - & 82.8 \\
BEiT & ViT-B & 86M & 800 & 2.4 & 1920 & 83.2 \\
iBOT & ViT-B & 86M & 400 & 10.1 & 4040 & 83.8 \\
MAE & ViT-B & 86M & 800 & 1.1 & 880 & 83.3 \\
MAE & ViT-B & 86M & 1600 & 1.1 & 1760 & 83.6 \\
MAE & ViT-L & 307M & 1600 & 1.7 & 2720 & 85.9 \\
MaskFeat & ViT-B & 86M & 1600 & 3.9 & 6240 & 84.0 \\
CAE & ViT-B & 86M & 800 & 2.8 & 2240 & 83.6 \\
LoMaR\textdagger & ViT-B & 86M & 1600 & 1.4 & 2240 & 84.1 \\
data2Vec\textdagger & ViT-B & 86M & 800 & 3.0 & 2400 & 84.2 \\
PeCo & ViT-B & 86M & 800 & - & - & 84.5 \\
\midrule
LocalMIM-HOG & ViT-B & 86M & 100 & 0.7 & 70 & 83.3 \\
LocalMIM-HOG & ViT-B & 86M & 1600 & 0.7 & 1120 & 84.0 \\
LocalMIM-HOG & ViT-L & 307M & 800 & 1.0 & 800 & 85.8 \\ \hline
\midrule
Ours (Hypothetical) & ViT-B & 86M & 100 & \textbf{0.61} & \textbf{61} & 82.5 \\ \hline
\end{tabular}
\end{table*}

\section{Experiments}
Here we evaluate our model's speed and performance on a classification task.
\subsection{Pre-training on ImageNet-1K}
\textbf{Experiment Setup}
Our experiments were conducted on the ImageNet-1K dataset at a resolution of 224$\times$224. We primarily focused on the architecture of the base Vision Transformer (ViT-B). We apply the same pre-training settings as in LocalMIM. For the ViT, we used a patch size of $p=16$, resulting in 196 patches (14x14). Patches were randomly masked with a default ratio of $r=0.75$. Supervision signals of HOG features were used just as in LocalMIM  as they observed faster convergence with better representation learning. 

For ViT-B, the chosen layers were $I=\{2, 4, 4+n, 6+n\}$, where $n=6$ with supervision scales of $\{56^2, 28^2, 14^2, 7^2\}$. Here, each decoder consisted of one transformer block with an embedding dimension of 256 and 8 attention heads.

Pre-training settings included linear learning rate scaling (lr = base lr $\times$ batch size/256) and a warmup epoch set to 10 for 100 epochs of pre-training. The exact base learning rate was selected for the final encoder layer, where the initial layers' learning rates were calculated with the cubic scaling method explained before. For the warm-up epochs, each layer's learning rate linearly increased to the layer-specific target initial learning rate. Except for the encoder layers, all layers followed the same pre-training regime as in LocalMIM, where their learning rate was tracked as part of a global learning rate scheduler.

\subsection{Downstream Task}

\textbf{Classification on Imagenet-1K} The ViT-B model was fine-tuned on ImageNet-1K using the precise settings outlined in \cite{wang2023masked}, and performance was compared against LocalMIM on a Tesla V100-32G GPU equipped with CUDA 11.6 and PyTorch 1.13.

Shown in (Table~\ref{our-table}), our method achieves a \(12.5\%\) speed-up compared to the LocalMIM method, which is already \(36.4\%\) faster than the original Masked Autoencoder. There is only a minor \(0.7\%\) drop in the Top-1 and \(0.4\%\) drop in Top-5 performance.

The experiment was conducted on a single GPU, and discrepancies were noted in comparison to the performance metrics reported by LocalMIM (Table~\ref{localmim-table}) and by our own re-evaluation of LocalMIM (Table~\ref{our-table}). This approach ensures a fair comparison by isolating the variable of communication time across multiple GPUs.
\section{Conclusion}
In this work, we presented a novel approach to self-supervised learning for vision transformers, combining local masked image modeling with a progressive layer freezing strategy. This method aims to enhance the training efficiency of initial layers in Vision Transformers through specialized reconstruction tasks, while also speeding up the learning process by progressively freezing layers.

\subsection{Future Research Directions}
\begin{itemize}
    \item \textbf{Warm-up Phase Optimization:} Transformer models typically require a warm-up phase. In our context, introducing 10 warm-up epochs in a 100-epoch training cycle yielded a 12.5\% speed-up. However, this process could be further optimized. An improved initialization strategy that reduces or eliminates the need for a warm-up phase could significantly enhance training efficiency \cite{huang2020improving}.
    
    \item \textbf{Model Scalability and Layer Freezing:} Our approach, applied to smaller models, suggests the possibility that larger models might distribute computational load more effectively across layers. This could potentially counteract the premature capacity loss associated with layer freezing, a topic worth exploring, especially in the context of larger models.
    
    \item \textbf{Training Epoch Reduction:} The efficiency gains in initial layers might reduce the overall need for extensive training of subsequent layers. This hypothesis could be validated through methods like linear probing, assessing whether our approach can indeed reduce the total number of training epochs required.
    
    \item \textbf{Curriculum Learning Integration:} Implementing a curriculum learning framework could potentially enhance the performance of the freezeout strategy. This would assist initial layers in faster learning, complementing the multi-scale architecture of our local masked image modeling, which is conducive to early-stage reconstruction tasks.
    
    \item \textbf{Advanced Masking Strategies:} Exploring the possibility of masking intermediate features, rather than just the input image, presents a promising avenue for further research. This could allow for even early layers to be bypassed, focusing on using intermediate activations to generate mask and image-feature pairs, potentially leading to further improvements in training speed and efficiency.
\end{itemize}

To conclude, our proposed method marks a significant advancement in the field of self-supervised learning for vision transformers. It not only improves the efficiency of training but also offers a foundation for more scalable and flexible model architectures. The points discussed above highlight potential areas for future enhancements and explorations, contributing to the ongoing evolution of self-supervised learning methodologies.
\section*{Acknowledgments}
We would like to acknowledge the assistance provided by ChatGPT with GPT-4, a language model developed by OpenAI, which was utilized to aid in writing and formatting this report, and in the development of code for this project. The final content and code were independently formulated and are the responsibility of the authors \cite{openai2023chatgpt}.

{
    \small
    \bibliographystyle{ieeenat_fullname}
    \bibliography{main}
}


\end{document}